\documentclass[letterpaper, 10 pt, conference]{ieeeconf}  % Comment this line out if you need a4paper

\IEEEoverridecommandlockouts                              % This command is only needed if 
\usepackage{amsmath} % assumes amsmath package installed
\usepackage{graphicx}
\usepackage{subfigure}
\usepackage{stfloats}
\usepackage{caption}
\usepackage{placeins} 
\usepackage{graphicx}
\usepackage{multicol}
\usepackage{flushend}
\usepackage{balance}
\usepackage{placeins}
\usepackage[absolute,overlay]{textpos}
\usepackage{wrapfig}
\usepackage{bm} 

\title{\LARGE \bf
Empowering Robot Teleoperation: Exploring the Synergies Between Devices and Manipulator Controllers with Quantitative analysis
}

\author{Yuxuan Zhao$^{2,3}$, Yuanchen Tang$^{1,3}$, Jindi Zhang$^{1,3,*}$, and Hongyu Yu$^{2,*}$% <-this % stops a space
\thanks{$^1$School of Science and Engineering, The Chinese University of Hong Kong, Shenzhen, Guangdong, China; $^2$Department of Mechanical and Aerospace Engineering, Hong Kong University of Science and Technology; $^3$Shenzhen Institute of Artificial Intelligence and Robotics for Society (AIRS), Shenzhen, Guangdong, China}
}

\begin{document}

\maketitle

\def\thefootnote{$^*$}\footnotetext{Corresponding Author}\def\thefootnote{\arabic{footnote}}
\thispagestyle{empty}
\pagestyle{empty}

%%%%%%%%%%%%%%%%%%%%%%%%%%%%%%%%%%%%%%%%%%%%%%%%%%%%%%%%%%%%%%%%%%%%%%%%%%%%%%%%
\begin{abstract}
Robot learning empowers the robot system with human brain-like intelligence to autonomously acquire and adapt skills through experience, enhancing flexibility and adaptability in various environments. Aimed at achieving a similar level of capability in large language models (LLMs) for embodied intelligence, data quality plays a crucial role in training a foundational model with diverse robot skills. In this study, we investigate the collection of data for manipulation tasks using teleoperation devices. Different devices yield varying effects when paired with corresponding controller strategies, including position-based inverse kinematic (IK) control, torque-based inverse dynamic (ID) control, and optimization-based compliant control. Analysis of experimental results suggests the importance of the relationship between teleoperation devices and controllers for real tasks.
\end{abstract}

%%%%%%%%%%%%%%%%%%%%%%%%%%%%%%%%%%%%%%%%%%%%%%%%%%%%%%%%%%%%%%%%%%%%%%%%%%%%%%%%
\section{INTRODUCTION}
Teleoperation technology \cite{ref1} has emerged as a critical enabler for robotic systems \cite{10196474}, particularly in high-risk environments where human intervention is either dangerous or impractical \cite{ref3}. By allowing operators to control robots remotely \cite{ref5}, teleoperation bridges the gap between human expertise and robotic precision, enabling complex tasks to be performed safely and efficiently \cite{ref6}. However, the effectiveness of teleoperation systems heavily depends on the integration of advanced devices\cite{Vision} and adaptive control strategies \cite{ref7}. Despite significant progress in this field, challenges such as latency \cite{ref8}, limited dexterity \cite{ref9}, and environmental adaptability \cite{ref10} continue to hinder the widespread adoption of teleoperation in critical applications, including power systems \cite{ref16}\cite{10902489}, disaster response, and industrial automation \cite{ref17}\cite{10930737}.

This study aims to address these challenges by investigating the optimal integration of teleoperation devices and control strategies to enhance robotic operational capabilities and task execution efficiency. Recent advancements in hardware, such as high-precision motion capture systems \cite{ref2} and haptic feedback devices \cite{ref13}, have expanded the possibilities for more intuitive and responsive teleoperation \cite{ref14}. Similarly, innovations in control algorithms, including high precision tracking and adaptive control techniques \cite{qp-multi-task}, have shown promise in improving system robustness and adaptability \cite{mr} \cite{hogan}. However, the synergistic effects of combining these devices and strategies remain underexplored, leaving a gap in understanding how to maximize their collective potential.

By analyzing the interplay between various teleoperation devices and control methods, this research seeks to provide guidance for robots in different tasks. Specifically, we focus on identifying configurations that minimize latency, enhance dexterity, and improve environmental adaptability, thereby enabling robots to perform complex tasks with greater precision and reliability. The findings of this study are expected to advance the application of teleoperation technology in high-risk operational scenarios, such as power grid maintenance, where replacing human workers with robots can significantly reduce safety risks and operational costs.

The paper is followed by several inter-connected sections
that present a comprehensive comparative study on data
collection for manipulation tasks. Section II gives a recent outlook for the robot teleoperation interface and the selected devices used in this study. Section III introduces the
preliminaries of robot modeling and the controller design
for manipulators to track the reference signal provided by
the human teleoperator. In section IV, the experiment set-
up is illustrated, and corresponding experimental results are
presented. Section V gives a supplementary discussion and
conclusion.

\section{Related Work}

Recent advancements in teleoperation and learning have led to diverse data collection methods, such as Stanford’s UMI \cite{umi}, Apple Vision Pro \cite{Apple}, and ALOHA \cite{Aloha}, each with unique strengths and limitations. In our work, we combined cameras and the Rokoko Smartsuit \cite{ROKOKO} to capture human arm motion data. Cameras provided precise skeletal capturing, while the Smartsuit offered full-body motion capture with IMUs. This hybrid approach ensured high-quality data collection, addressing limitations like occlusions or calibration challenges, and enabled robust datasets for teleoperation and human-robot interaction.

In the teleoperation data collection for dexterous hands, we primarily employ the following two methods: WiLoR and Rokoko Smartgloves. WiLoR \cite{WILoR} is an end-to-end 3D hand localization and reconstruction method, specifically designed for multi-hand detection and reconstruction in complex environments, such as varying lighting, occlusion, and motion blur. It enhances model robustness through a large-scale dataset, WHIM, and supports smooth 3D hand tracking in monocular videos. Rokoko Smartgloves, on the other hand, are intelligent gloves designed for motion capture, accurately capturing subtle hand and finger movements through built-in sensors and providing high-precision real-time data feedback.

This study employs cameras to capture the motion data of the exoskeleton robotic arm, integrated with the WiLoR system for hand motion tracking. Simultaneously, the Rokoko SmartSuit Pro II is utilized to collect arm motion data, complemented by Rokoko Smartgloves for hand motion acquisition, achieving synchronized collection of arm and hand motion information.

\section{Controller Design}
\subsection{Prelinminaries} \label{sec:pre}
Robot comprises several linkages that are connected by different types of actuated/passive joints.
The motion of the robot could be described by all actuated joints in a position-/torque- controlled manner. In \cite{mr}, it introduces the forward kinematic relationship between joint and task space, which could be written as
\begin{equation}
    \dot{\textit{\textbf{x}}} = \textit{\textbf{J}}(\textit{\textbf{q}})\dot{\textit{\textbf{q}}}
    \label{eq.1}
\end{equation}
where $\textit{\textbf{q}} \in R^n$ is the generalized joint coordinate, $x \in R^m$ is a vector of task variable, and $\textit{\textbf{J}}(\textit{\textbf{q}}) = \frac{\partial{\dot{\textit{\textbf{x}}}}}{\partial{\dot{\textit{\textbf{q}}}}} \in R^{m\,\text{x}\,n}$ is the Jacobian matrix.

Jacobian matrix builds the connection between the motion in the robot joint space and the Cartesian space. By taking the derivative of the equation (\ref{eq.1}), the task acceleration $\ddot{x}$ could be derived as below
\begin{equation}
    \ddot{\textit{\textbf{x}}} = \textit{\textbf{J}}(\textit{\textbf{q}})\ddot{\textit{\textbf{q}}} + \dot{\textit{\textbf{J}}(\textit{\textbf{q}})}\dot{\textit{\textbf{q}}}
\end{equation}

Given the Jacobian matrix, the manipulator dynamic model with external interaction could be expressed as 
\begin{equation}
     \textit{\textbf{M}}(\textit{\textbf{q}})\ddot{\textit{\textbf{q}}} + \textit{\textbf{C}}(\textit{\textbf{q}},\dot{\textit{\textbf{q}}})\dot{\textit{\textbf{q}}} + \textit{\textbf{G}}(\textit{\textbf{q}}) = \bm{\mathit{\tau}} + \textit{\textbf{J}}^T(\textit{\textbf{q}})\textit{\textbf{F}}_{ext}
    \label{eq:dynamic}
\end{equation}
with the positive definite inertial matrix $\textit{\textbf{M}}(\textit{\textbf{q}}) \in R^{n \, \text{x} \, n}$, and the Coriolis and centrifugal matrix $\textit{\textbf{C}}(\textit{\textbf{q}},\dot{\textit{\textbf{q}}}) \in R^{n \, \text{x} \, n}$. The term $\textit{\textbf{G}}(\textit{\textbf{q}}) \in R^n$ represents the effect due to the generalized gravity force. Moreover, $\bm{\mathit{\tau}} \in R^n, \ \textit{\textbf{F}}_{ext}\in R^{m}$ are generalized joint torque, and the external wrench is exerted in the robot respectively.
\begin{figure*}[ht]
    \centering
    \includegraphics[width=0.8\linewidth]{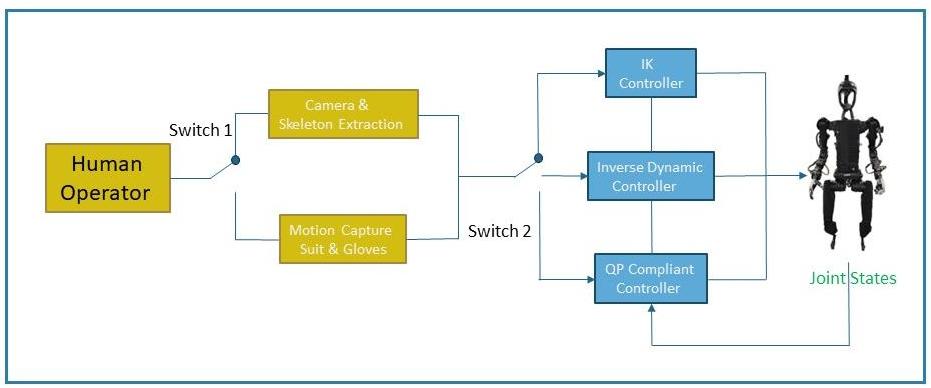}
    \caption{Framework of the teleoperation process for data collection}
    \label{fig:framework}
\end{figure*}
During the teleoperation period, the combination of different interfaces and the controller design will lead to a different performance in the accuracy of the human operator's target and controller tracking. The entire comparative study, including the category of the interface and the corresponding controller, could be depicted in the Figure \ref{fig:framework}. The remaining of the section will provide more details about the different control approaches for the robot to track the human teleoperator's command by utilizing the robotics background knowledge in Section \ref{sec:pre}.

\subsection{Position- and Torque-based Controller} \label{sec:posi/torque}
Given the kinematic mapping by the Jacobian matrix, the desired joint configuration could be easily computed by the inverse of the Jacobian matrix. However, when the Jacobian matrix is not full rank, it means the robot is in a singular situation. The singularity could occur due to the lack of degree of freedom (D.O.F) relative to the task space dimension or a specific joint configuration in which the robot lost one or more degrees of freedom in certain Cartesian directions. The existence of the singularity is unfavorable because the inverse of the Jacobian matrix will give an extremely large quantity for the joint velocity, resulting in unstable behavior and leading to damage to the robot hardware. On the other hand, the redundancy of the robot D.O.F. offers multiple solutions to the commanded joint velocity. To mitigate both issues, a damped least squares method is utilized. The pseudo-inverse of the Jacobian is first computed to minimize the 2-norm of the joint velocity while tracking the desired task configuration. When the robot is approaching the singularity, a small positive damping term $\lambda$ is applied to make the robot pass through the singularity point. The damped least squares inverse kinematic approach could be formulated as the following 
\begin{equation}
    \dot{\textit{\textbf{q}}}_{des} = \begin{cases}
       \textit{\textbf{J}}^T(\textit{\textbf{q}})(\textit{\textbf{J}}(\textit{\textbf{q}})\textit{\textbf{J}}^T(\textit{\textbf{q}}))^{-1}\dot{\textit{\textbf{x}}}_{des}, & \delta(\textit{\textbf{q}}) > \epsilon\\
       \textit{\textbf{J}}^T(\textit{\textbf{q}})(\textit{\textbf{J}}(\textit{\textbf{q}})\textit{\textbf{J}}^T(\textit{\textbf{q}}) + \lambda \textit{\textbf{I}})^{-1}\dot{\textit{\textbf{x}}}_{des}, & \delta(\textit{\textbf{q}}) \leq \epsilon
    \end{cases}
    \label{dls}
\end{equation}
where $\delta(\textit{\textbf{q}}) = \sqrt{\text{det}(\textbf{\textit{J}}\textbf{\textit{J}}^T)}$ represents the manipulability function of the robot arm, and $\epsilon$ is a small positive threshold to determine whether the robot is at the singular point. The desired twist $\dot{\textit{\textbf{x}}}_{des}$ could be computed by numerical differentiation based on the desired 6D pose and sampling time. The desired position for joint variables could be obtained by numerical integration from the desired joint velocity calculated by equation (\ref{dls}).

Another control strategy, the torque-based inverse dynamic controller, attempts to enforce the robotic dynamics by commanding the desired torque to the robot's joints. By considering the robot state, including generalized joint position and velocity, we could convert the desired task pose to the desired joint velocity in the next timestep using the scheme of inverse kinematics. Additionally, the desired generalized joint acceleration could be obtained via differentiation of the desired joint velocity. Thus, the joint acceleration and velocity could be converted to the joint torques governed by equation (\ref{eq:dynamic}).

\subsection{QP Compliant Controller}
In this section, we introduce the quartic programming-based compliant controller derived from the task-space QP control formulation in \cite{qp-multi-task}. The general whole-body controller forms an optimization problem to minimize the weighted sum of the task costs corresponding to different objectives. Simultaneously, the constraints of the optimization problem modify the QP search space to accommodate the hardware limitations, dynamic feasibility, and extra physical enforcement.

The main objective is to track the task space goal and render the end-effector of the arm as a compliant model such as the mass-spring-damper, or mass-damper dynamics. This concept of compliant control is first introduced by Hogan in \cite{hogan}, and he attempts to implement a dynamic relation during the interaction between the robot and environment. We introduce impedance tracking task formulation to achieve the impedance controller design ideology. During teleoperation, the controller needs to track the reference pose $x_{des}$ and twist $\dot{x}_{des}$. The reference signals are used to calculate the dynamic wrench residue, resulting in
\begin{equation}
    \textit{\textbf{f}}_{des} = \textit{\textbf{K}}_{des}(\textit{\textbf{x}}_{des} - \textit{\textbf{x}}) + \textit{\textbf{D}}_{des}(\dot{\textit{\textbf{x}}}_{des} - \textit{\textbf{J}}(\textit{\textbf{q}})\dot{\textit{\textbf{q}}})
\end{equation}
where the task posture error is regulated by the translation difference and quaternion difference for orientation. The $\textit{\textbf{K}}_{des}$ is the user-specified 6 x 6 positive definite diagonal stiffness matrix. The diagonal entries express the corresponding Cartesian stiffness for the manipulator end-effector. \cite{close-loop-damping} proved that the closed-loop critical damping system could be achieved by setting 
\begin{equation}
    \textit{\textbf{D}}_{des} = \mathit{\mathbf{\Lambda}}^{\frac{1}{2}}\textit{\textbf{K}}^{\frac{1}{2}}_{des} + \textit{\textbf{K}}^{\frac{1}{2}}_{des}\Lambda^{\frac{1}{2}}
\end{equation}
where $\mathit{\mathbf{\Lambda}} \in R^{m \, \text{x} \, m}$ is the task-space inertial matrix and expressed as 
\begin{equation}
    \mathit{\mathbf{\Lambda}} = (\textit{\textbf{J}}(\textit{\textbf{q}})\textit{\textbf{M}}(\textit{\textbf{q}})^{-1}\textit{\textbf{J}}(\textit{\textbf{q}})^T)^{-1}
\end{equation}
Since the $\textit{\textbf{K}}_{des}$ is a diagonal matrix, the square root of the stiffness matrix could be calculated by element-wise square root operation. The square root of the task-space inertial matrix needs to be calculated via Eigendecomposition. The factorized eigenvectors $\textit{\textbf{P}}$ and diagonal eigenvalue matrix $\textit{\textbf{D}}$ could compute the square root of the ask-space inertial matrix as follows:
\begin{equation}
    \mathit{\mathbf{\Lambda}}^{\frac{1}{2}} = \textit{\textbf{PD}}^{\frac{1}{2}}\textit{\textbf{P}}^T
\end{equation}
Furthermore, the track objective error function of QP could be formulated as 
\begin{equation}
    \textit{\textbf{e}}_{track} = \ddot{\textit{\textbf{x}}}_{des} - \mathit{\mathbf{\Lambda}}^{-1}\textit{\textbf{f}}_{des}
\end{equation}
With the sole impedance tracking objective, the singularity problem mentioned in Section\ref{sec:posi/torque} will make the hessian matrix of the QP problem non-positive definite, resulting in unstable control behavior. To address the problem, we introduce another objective to regulate the nominal position for joint configuration that is away from the singular point. A diagonal selection matrix $\textit{\textbf{S}} \in R^{n \, \text{x} \, n}$ contains only zeros and ones to decide the activating joints. The so-called null space joint PD error forms the joints' acceleration as 
\begin{equation}
    \ddot{\textit{\textbf{q}}}_{feedback} = \textit{\textbf{K}}_{n}(\textit{\textbf{q}}_{des} - \textit{\textbf{q}}) + \textit{\textbf{D}}_n(\dot{\textit{\textbf{q}}}_{des} - \dot{\textit{\textbf{q}}}) 
\end{equation}
with user-defined diagonal proportion and derivative gain matrices $\textit{\textbf{K}}_n, \, \textit{\textbf{D}}_n \in R^{n \, \text{x} \, n}$. Hence, the corresponding joint null space error function is defined by
\begin{equation}
    \textit{\textbf{e}}_{joint} = \textit{\textbf{S}}(\ddot{\textit{\textbf{q}}}_{des} - \ddot{\textit{\textbf{q}}}_{feedback})
\end{equation}

Based on the formulations of the previous paragraph, the final QP problem for the compliant force controller could be constructed as 
\begin{gather*} 
    \ddot{\textit{\textbf{q}}}_{des} = \underset{\ddot{\textit{\textbf{q}}}_{des}}{\text{arg min}} \; \textit{\textbf{e}}_{track}^T\textit{\textbf{W}}_{track}\textit{\textbf{e}}_{track} + \textit{\textbf{e}}_{joint}^T\textit{\textbf{W}}_{joint}\textit{\textbf{e}}_{joint}\text{,} \\
    \text{s.t. joint position, velocity, torque limitations (\ref{c:posi}), (\ref{c:vel}), (\ref{c:tau})}
    \label{qp_formu}
\end{gather*}
The hardware limitation will enforce the constraints for joint position, velocity, and torque. Inspired by \cite{qp-multi-task}, we formulated the constraints as following 
\begin{gather}
   \textit{\textbf{q}}_{min} - \dot{\textit{\textbf{q}}}\Delta t - \textit{\textbf{q}} \leq \frac{1}{2}\Delta t^2\ddot{\textit{\textbf{q}}} \leq \textit{\textbf{q}}_{max} - \dot{\textit{\textbf{q}}}\Delta t - \textit{\textbf{q}}  \label{c:posi}\\
   \dot{\textit{\textbf{q}}}_{min} - \dot{\textit{\textbf{q}}} \leq \Delta t\ddot{\textit{\textbf{q}}} \leq \dot{\textit{\textbf{q}}}_{max} - \dot{\textit{\textbf{q}}} \label{c:vel} \\
   \bm{\mathit{\tau}}_{min} \leq  \textit{\textbf{M}}(\textit{\textbf{q}})\ddot{\textit{\textbf{q}}} + \textit{\textbf{C}}(\textit{\textbf{q}},\dot{\textit{\textbf{q}}})\dot{\textit{\textbf{q}}} + \textit{\textbf{G}}(\textit{\textbf{q}})\leq \bm{\mathit{\tau}}_{max} \label{c:tau}
\end{gather}
From the optimization problem in \ref{qp_formu}, the decision variable, joint acceleration will be searched in the constrained space and is used as the desired joint acceleration $\ddot{q}_{des}$. To actuate the robot in a torque-controlled way and enforce the robot dynamic model, the desired joint acceleration is converted to the joint torque by $\tau_{des} = \textit{\textbf{M}}(\textit{\textbf{q}})\ddot{\textit{\textbf{q}}}_{des} + \textit{\textbf{C}}(\textit{\textbf{q}},\dot{\textit{\textbf{q}}})\dot{\textit{\textbf{q}}} + \textit{\textbf{G}}(\textit{\textbf{q}})$.

\section{Experiment}
\begin{figure}[ht]
    \centering
    \includegraphics[width=0.85\linewidth]{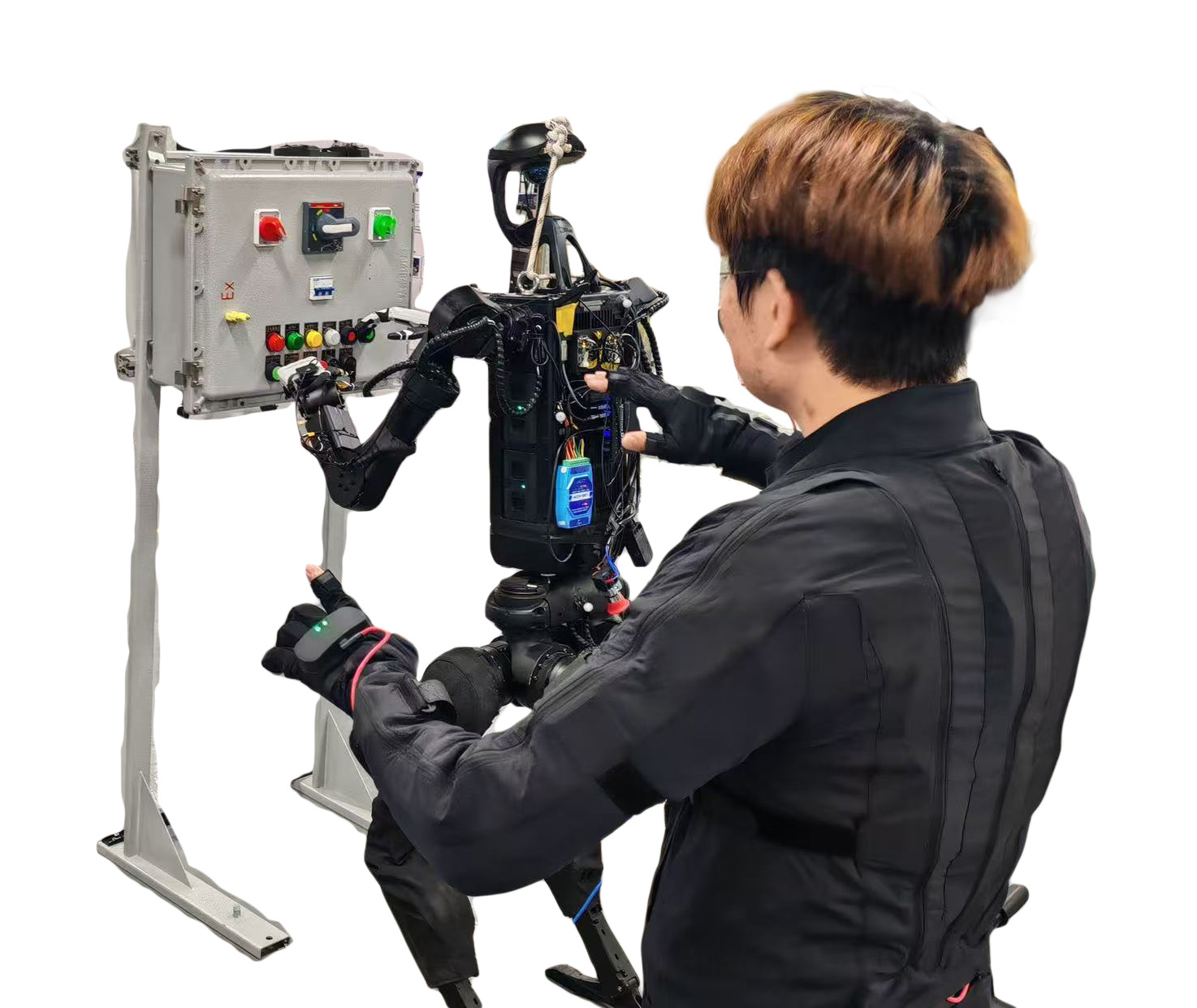}
    \caption{A human operator wearing the motion capture suit teleoperates the humanoid robot to press the buttons of the electrical box.}
    \label{fig:tele}
\end{figure}
\subsection{Experiment Setup and Implementation Detail}
\begin{figure}[ht]
    \centering
    \subfigure[WiLoR]{
        \includegraphics[width=0.52\linewidth]{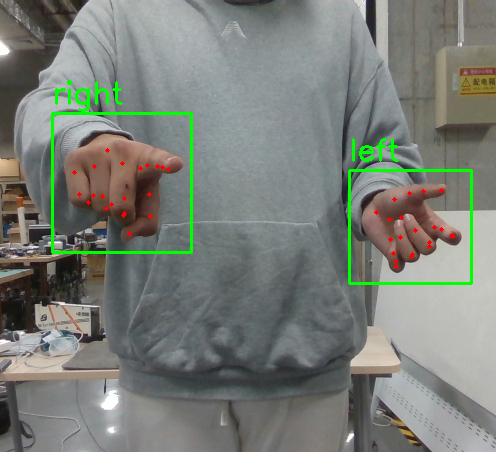}
        \label{fig:hand}
    }
    \subfigure[Human exoskeleton]{
        \includegraphics[width=0.36\linewidth]{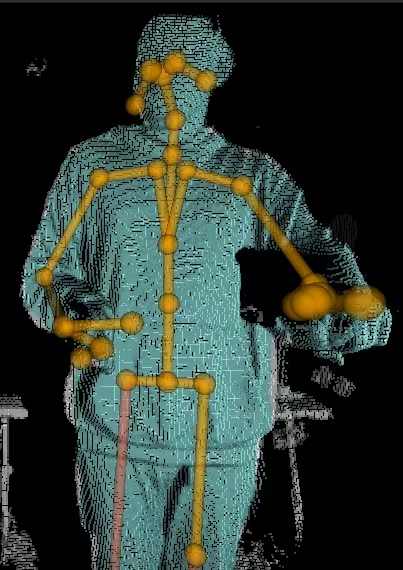}
        \label{fig:skeleton}
    }
    \caption{Camera extracts human exoskeleton and collaborates with WilorR to extract hand information.}
    \label{fig:both}
\end{figure}

This study is conducted with the Unitree Robotics H1 humanoid robot system with the focus of the upper body manipulation task for teleoperated data collection for specific power grid scenario, as shown in Figure \ref{fig:tele}. To explore the synergies within the teleoperation process, we redesigned the forearms of H1 robot to assign three more degrees of freedom in its arms. Hence the robot comprises two 7 D.O.F. arms to manipulate an electrical box containing buttons, circuit breakers, and rotational switches. In terms of the teleoperation interface, the human skeleton was extracted using the Azure Kinect body tracking SDK\footnote{Body tracking SDK: https://packages.microsoft.com} and demonstrated in Figure \ref{fig:skeleton}. By leveraging the human skeleton data, the positions of the human's hands were mapped to the end effectors' positions, expressed in the robot's torso frame. Alongside the corresponding RGB image, we adopted the WiLoR method for hand pose estimation, as shown in Figure \ref{fig:hand}, providing orientation targets represented in unit quaternion for the controllers to track.

In addition to the camera interface, the motion capture suit facilitated direct tracking of the end effector targets from the human operator, allowing for precise adjustments of the robot's physical measurements, such as height, forearm length, shoulder width, and similar parameters. For the implementation of three controllers, we utilized RBDL \cite{RBDL} for robot kinematics and dynamics modeling and the C++ Eigen interface of the OSQP \cite{osqp} solver named osqp-eigen. The parameters for the proposed QP controller are illustrated in the Table \ref{tab:parameter}.

\begin{table}[!htbp]
    \centering
    \scalebox{1.5}{
        \begin{tabular}{|c|c|}
            \hline
            Paramter & Value \\
            \hline
             $\textit{\textbf{K}}_n$ & diag(80, 80, 80, 5, 5, 5) \\
             \textit{\textbf{S}} & diag(1, 1, 0, 0, 0, 0, 0) \\
             $\textit{\textbf{W}}_{track}$ & $15I_6$ \\
             $\textit{\textbf{W}}_{joint}$ & $1I_n$ \\
             $\Delta_T$ & 0.001\\
             \hline
        \end{tabular}
    }
    \caption{Table containing the parameters for the proposed QP controller}
    \label{tab:parameter}
\end{table}

The usage of the interface is determined by switch 1 of Figure \ref{fig:framework}, while switch 2 controls the respective controllers employed for tracking the human operator's target positions.

\subsection{Comparison between Interfaces}
\begin{figure}[!htb]
    % \centering
    \hspace{-0.02\linewidth}\includegraphics[width=1.05\linewidth]{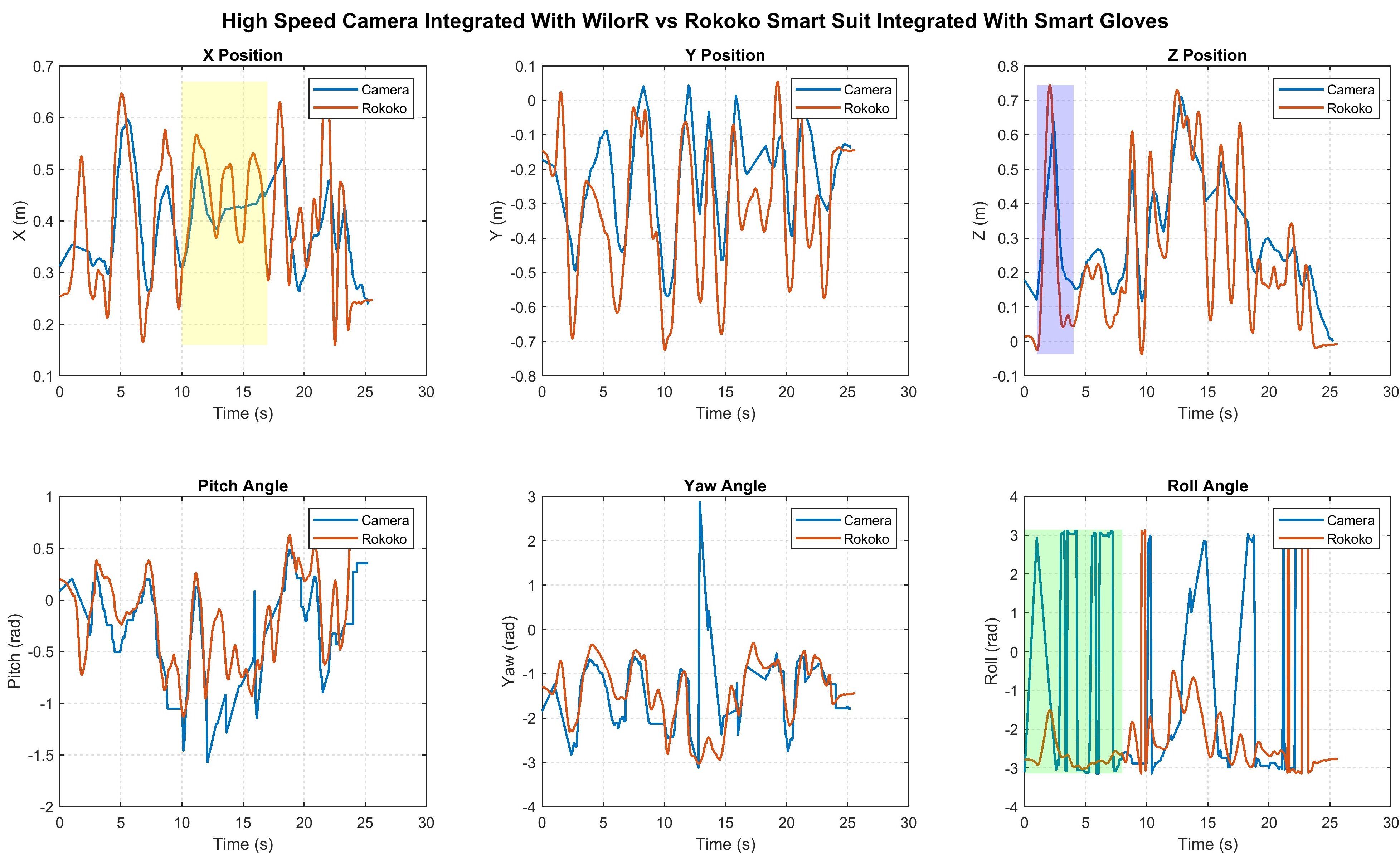}
    \caption{Camera Integrated with WiLoR vs Rokoko Smart Suit Integrated with Smart Gloves}
    \label{fig:target}
\end{figure}
To verify the quality of two types of teleoperation data acquisition, we had the operator perform the same action during the same time period, recording the data using a camera paired with WiLorR, as well as Rokoko Smartsuit paired with Rokoko Smartgloves. 

The results, as shown in Figure \ref{fig:target}, indicate that the combination of Rokoko smartsuit and Rokoko smartgloves outperforms the combination of camera and WilorR in terms of the completeness and accuracy of motion information capture, although there is a certain degree of data loss. In the blue area of Figure \ref{fig:target}, the peak of the red line (Rokoko system) appears earlier than that of the blue line (camera system), reflecting the superior real-time performance of the Rokoko system. Notably, in the green area, the camera and WiLorR combination is prone to misreading at $+\pi$ and $-\pi$, which may lead to significant posture fluctuations. Such misjudgment of posture information could potentially cause excessive motor current, abnormal speed, and other failure risks, posing potential damage to the equipment.

\subsection{Comparison between Controllers}
\begin{figure}[bp]
    % \centering
    \hspace{-0.02  \linewidth}\includegraphics[width=1.05\linewidth]{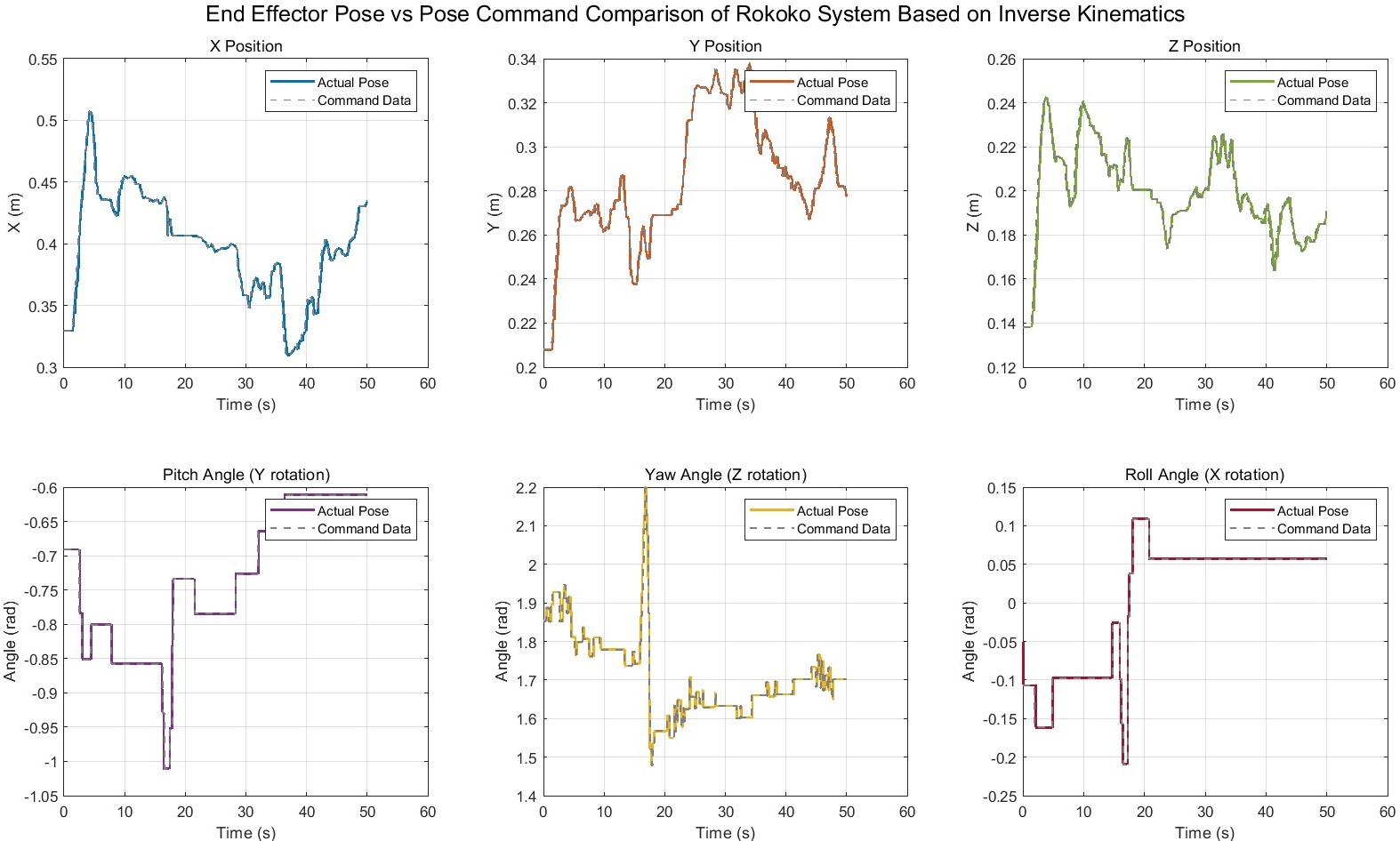}
    \caption{End Effector Pose vs Pose Command Comparison of Camera System Based on Inverse Kinematics}
    \label{fig:ik}
\end{figure}

\begin{figure}[!bp]
    % \centering
    \hspace{-0.02\linewidth}\includegraphics[width=1.05\linewidth]{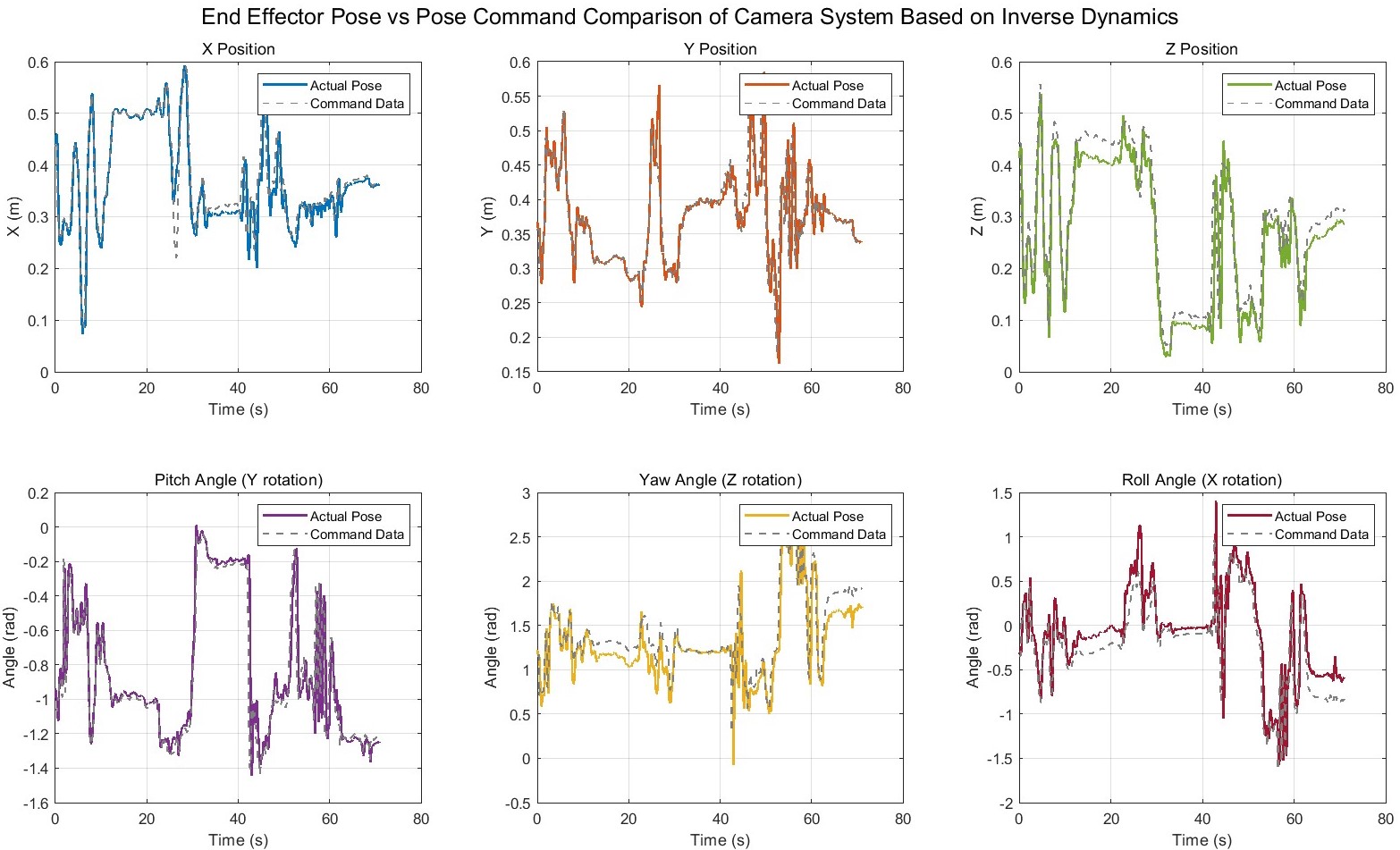}
    \caption{End Effector Pose vs Pose Command Comparison of Camera System Based on Inverse Dynamics}
    \label{fig:id}
\end{figure}

\begin{figure}[tp]
    % \centering
    \hspace{-0.02\linewidth}\includegraphics[width=1.05\linewidth]{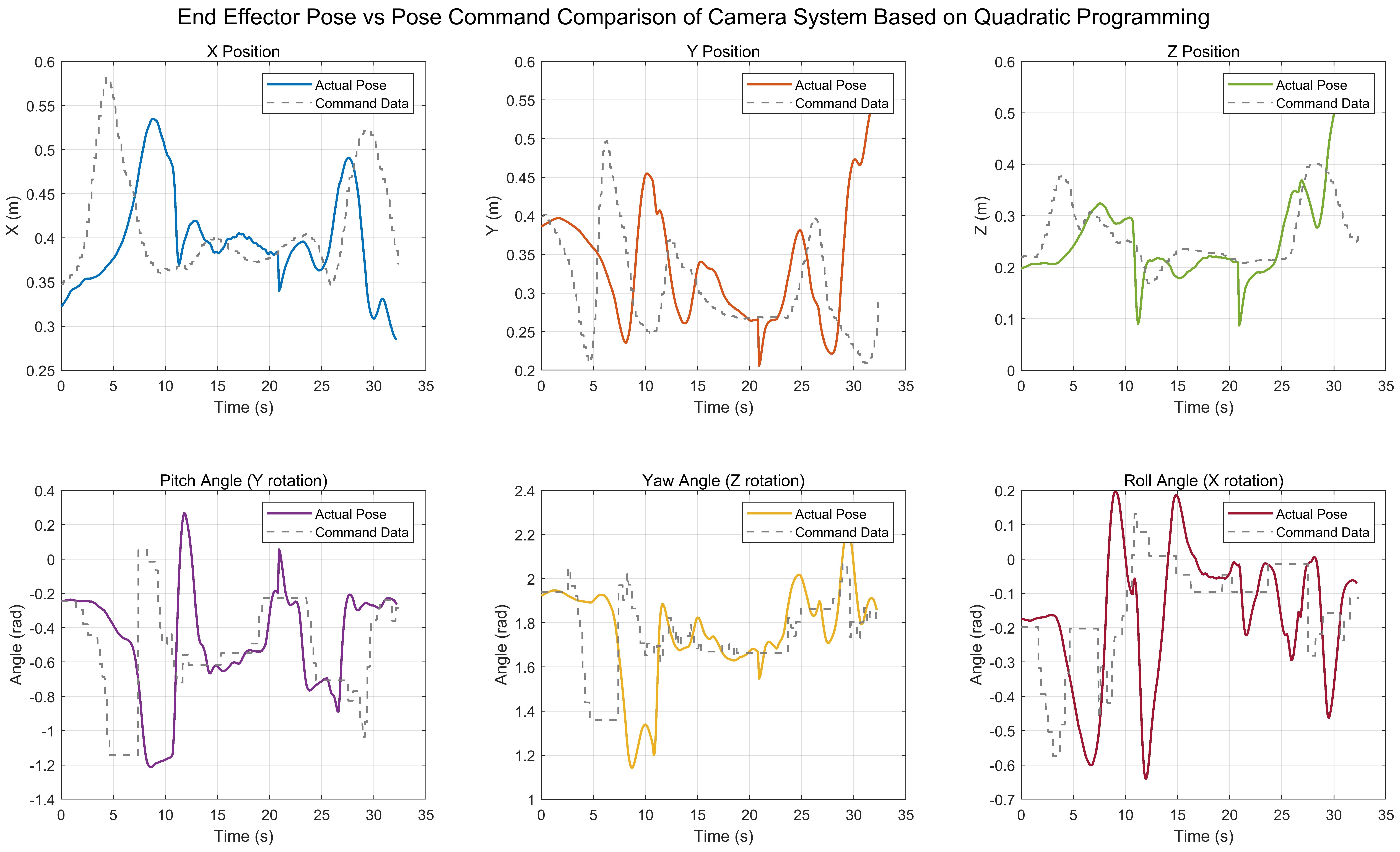}
    \caption{End Effector Pose vs Pose Command Comparison of Camera System Based on Quadratic Programming}
    \label{fig:qp}
\end{figure}

The Figures \ref{fig:ik}, \ref{fig:id}, \ref{fig:qp} present the data acquisition results of the same teleoperation device (Camera system) under different control methods (Inverse Kinematics, Inverse Dynamics, Quadratic Programming).

From Figure \ref{fig:ik}, we can observe that the end-effector of the arm strictly follows the teleoperated target with the position error accuracy lower than  6.20\% and the orientation error accuracy lower than 4.48\%. All the tracking error accuracies are defined as follows:
\begin{equation}
    \eta=\frac{1}{n}\sum_{i=1}^n \frac{|error_i|}{target_i}
\end{equation}
where the $|error_i|$ is the maximum absolute error while tracking the target variable. Even though the IK controller could provide an accurate tracking performance, sometimes the joint velocity might have a sudden change due to the infinite solution of the redundant robotic arm. The Figure \ref{fig:id} shows a poor tracking performance relative to the IK controller but still have the error accuracy lower than 13\%. The reason might be attributed to the mismatch of the robot dynamic model. Simultanously, the ID controller introduces a certain level of compliance while the robot is interacting with the external environment. The QP controller shows the worst tracking performance from the Figure \ref{fig:qp}. Moreover, we can spot there exists a lagging during tracking because of the limit of the joint velocity. On the other hand, the tracking pose trajectory becomes relatively smoother than the command, which is the benefit of the constraints (\ref{c:posi}), (\ref{c:vel}) formulated in the QP problem.

Through comparative analysis, it is evident that the control method based on inverse kinematics offers higher control precision, while the method based on quadratic programming demonstrates superior compliance. The inverse kinematics-based approach achieves a commendable balance between control precision and compliance. 

Consequently, in practical applications, for tasks demanding high control precision (such as precision manufacturing machinery and optical component calibration), it is advisable to employ the control method based on inverse kinematics. Conversely, for tasks requiring high compliance (such as human-robot collaborative assembly in factories), the QP-based control method is more suitable. For scenarios necessitating a balance between precision and compliance, such as nuclear waste handling (where precise grasping of hazardous materials and collision prevention safety are both critical), the control method based on inverse dynamics is more appropriate.

\section{CONCLUSIONS}
This paper presents a comparative study on the robot teleoperation data collection for the specific power grid scenario based on different teleoperation devices and the control strategy. First, the Rokoko system demonstrates significant advantages compared to the Camera system, specifically in terms of improved real-time performance, increased data completeness, and enhanced pose calibration accuracy. While the camera system may suffer from the problem of accuracy and real-time performance, it offers a more cost-effective solution. Additionally, the camera system can be easily deployed in various scenarios, including outdoor environments. Second, comparative studies of control methods for the same teleoperation device reveal a clear gradient in performance differences among the three control strategies: inverse kinematics, inverse dynamics, and multi-objectives quadratic programming. In terms of compliance, these methods show a progressive improvement, while their trajectory tracking accuracy correspondingly decreases, forming a typical trade-off relationship in control characteristics.

\addtolength{\textheight}{-12cm}   % This command serves to balance the column lengths

%%%%%%%%%%%%%%%%%%%%%%%%%%%%%%%%%%%%%%%%%%%%%%%%%%%%%%%%%%%%%%%%%%%%%%%%%%%%%%%%

%%%%%%%%%%%%%%%%%%%%%%%%%%%%%%%%%%%%%%%%%%%%%%%%%%%%%%%%%%%%%%%%%%%%%%%%%%%%%%%%


\section*{ACKNOWLEDGMENT}
The authors would like to thank Mr. Chongshan He for his countless hours of help and support for experiment setup and inspiring discussion.

% This work is supported by the

%%%%%%%%%%%%%%%%%%%%%%%%%%%%%%%%%%%%%%%%%%%%%%%%%%%%%%%%%%%%%%%%%%%%%%%%%%%%%%%%

\bibliographystyle{ieeetr}
\bibliography{reference}

\vfill
\end{document}